\documentclass{article}

\usepackage[final,nonatbib]{nips_2018}

\usepackage[utf8x]{inputenc} 
\usepackage[T1]{fontenc}    
\usepackage{hyperref}       
\usepackage{url}            
\usepackage{booktabs}       
\usepackage{amsfonts}       
\usepackage{nicefrac}       
\usepackage{microtype}      
\usepackage{float}
\usepackage[hang,flushmargin]{footmisc}
\usepackage{graphicx}
\usepackage{subfigure}
\usepackage{amssymb}
\usepackage{mathtools}
\usepackage{booktabs}
\usepackage{enumitem}
\usepackage{makecell}
\usepackage{pbox}
\usepackage{hhline}

\usepackage{color}

\title{Privacy-Preserving Action Recognition for Smart Hospitals using Low-Resolution Depth Images}

\author{
Edward Chou$^1$ \qquad
Matthew Tan$^1$ \qquad
Cherry Zou$^1$ \qquad
Michelle Guo$^1$ \qquad \\
\textbf{Albert Haque$^1$ \qquad
Arnold Milstein$^2$ \qquad
Li Fei-Fei$^1$} \\
\\
$^1$Department of Computer Science, Stanford University\\
$^2$Clinical Excellence Research Center, School of Medicine, Stanford University
}

\begin{document}

\maketitle

\begin{abstract}
Computer-vision hospital systems can greatly assist healthcare workers and improve medical facility treatment, but often face patient resistance due to the perceived intrusiveness and violation of privacy associated with visual surveillance. We downsample video frames to extremely low resolutions to degrade private information from surveillance videos.  We measure the amount of activity-recognition information retained in low resolution depth images, and also apply a privately-trained DCSCN super-resolution model to enhance the utility of our images.  We implement our techniques with two actual healthcare-surveillance scenarios, hand-hygiene compliance and ICU activity-logging, and show that our privacy-preserving techniques preserve enough information for realistic healthcare tasks.
\end{abstract}


\section{Introduction}\label{sec:intro}
Healthcare facilities and services can benefit greatly from automation and machine vision to improve patient care and outcomes. A new paradigm commonly known as “smart hospitals” \cite{yu2018smarthospitaliot, noury2008smarthospitaldaynight, biswas2006smarthospitalsystem, twinanda2015smarthospitaldatadriven} aim to integrate automation and machine intelligence directly into the healthcare environment by using sensor-collected data to understand and facilitate hospital procedures. A promising approach leverages the information richness of visual data, using cameras and computer vision to collect and analyze patient and healthcare worker activities \cite{sanchez2008smarthospitalactivityrecognition}.  Vision-based activity monitoring has been deployed to track hand-hygiene compliance \cite{haque2017handhygiene}, perform activity logging in ICU facilities \cite{ma2017icumobility, liu2018icu}, and detect anomalous behaviors like falls in senior home facilities \cite{luo2018seniorhome}, demonstrating the potential of such systems to reduce disease, decrease human workload, and improve patient care.

Although smart hospitals result in better healthcare services, they can also elicit distrust from patients and healthcare workers \cite{lin2016iotsmarthomeprivacy}. A building filled with cameras performing constant monitoring can appear intrusive and oppressive, and the medical field as a whole is especially concerned with patient and data privacy \cite{asghar2017hipaaprivacyhealth}.  A visual system designed to enforce hand-hygiene compliance will invariably capture auxiliary information which could contain a patient’s identity, their medical condition, and personally-embarrassing activities.  Though using different data-modalities with depth sensors can alleviate some privacy concerns by avoiding RGB data \cite{zhang2012privacyfalldetection}, previous works show that common depth sensors like the Kinect capture enough data to perform facial recognition \cite{cheng2017facekinectdepth}.

One approach for alleviating patient concern over camera intrusiveness is to ensure that cameras capture as little privacy-sensitive information as possible.  In this work, we use low-resolution depth images to remove privacy-relevant information while still retaining activity-recognition utility.  We train deep learning models to perform hand-hygiene monitoring and activity-logging tasks and measure the accuracy drop due to downsampling the depth data. We enhance the utility of our images by using super-resolution techniques trained on a privacy-safe data to enhance our downsampled images, demonstrating a realistic framework for non-intrusive healthcare monitoring.


\section{Related Work}

The use of extremely low resolution images to perform activity recognition in a privacy-preserving manner has been explored in previous literature. One work \cite{miyazaki2015lowresprivacyconscious} proposes low-resolution action recognition by focusing on the shape of the human head to guide body-position estimation. Inverse super-resolution (ISR) \cite{ryoo2016lowresegocentric} uses a network generates multiple low resolution proposals and applies MCMC and entropy-measure approaches to discover the optimal transformation for action recognition.
Two similar approaches \cite{ryoo2017lowressiamese, xu2018lowresactiontwostream} use two stream neural networks to aggregrate features and create a cross representation between high and low resolution images to learn an optimal feature mapping.

Several works also explore low-resolution facial recognition which could be applicable to a privacy-sensitive context.  One such work \cite{li2018lowresfacewild} attempts to learn a common feature space between low and high resolution images using center regularization and GAN-based techniques.  Another \cite{zangeneh2017lowresfacetwobranch} uses a two branch network to learn a cross representation between low and high-resolution faces.

Enhancing the quality of low-resolution images is also a well-explored task, with many recent works focused especially on applying deep neural networks for superresolution \cite{yang2018superresdeeplearning}.  An effective approach \cite{yamanaka2017superresskipconnection, zhang2018superresresidualattention} uses skip/residual connections to bypass abundant low frequency information and focus the model on high frequency information for training.  One work focuses instead on compacting networks and introducing a compression component to perform superresolution tasks with computational improvements \cite{hui2018superresinfodistillation}.  Another distinct approach uses adversarial training techniques to generate realistic textures and applies the technique to video data \cite{perezpellitero2018superrresphotorealistic}.

\begin{figure}[t]
\centering
\vspace{-5mm}
\includegraphics[width=0.9\linewidth]{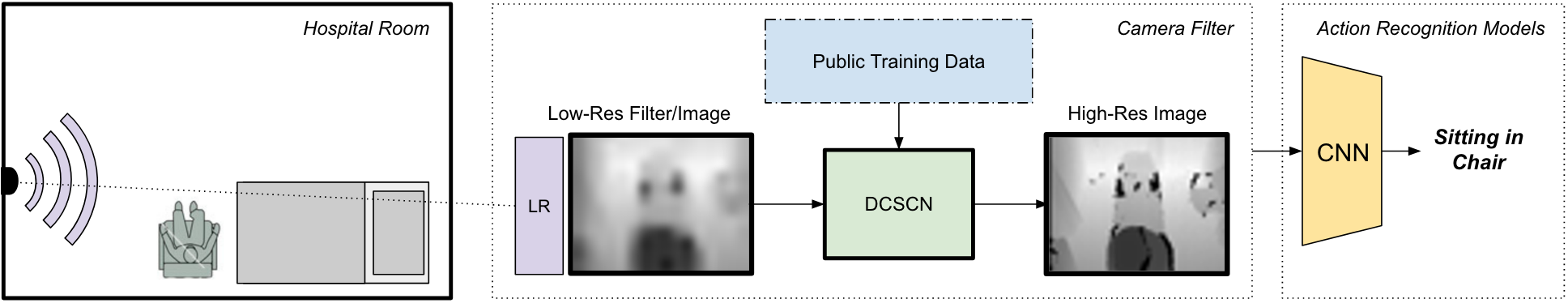}
\caption{A low-resolution camera is used to monitor a hospital room.  Public training data is used to train a super-resolution model to enhance images before fed to an action recognition model.}
\label{fig:method}
\end{figure}

\section{Methods}
A framework of our method is illustrated in Figure \ref{fig:method}. Low-resolution videos of a hospital room are captured and enhanced with a privacy-preserving DCSCN model, before the frames are fed to an action recognition model to perform tasks such as hand-hygiene monitoring or activity logging. 

\textbf{Downsampling:} In order to simulate image downsampling, we used bicubic downsampling of the original $224 \times 224$ depth images by different scales.  Other methods to distort images include Gaussian blurring or superpixel clustering \cite{butler2015privacyutilityrobots}. Bicubic interpolation downscaling works well with images that have continuous-tone images \cite{keys1981bicubic}, suitable for depth images that do not have many sharp edges.  We note that we can avoid collecting high-fidelity images altogether by using LR camera hardware \cite{ryoo2016lowresegocentric}, which is used in other settings to reduce memory for data storage. It has been found in previous work that at image sizes of ${\sim}  15 \times {\sim} 15$, there is not enough visual information to discern facial features, providing a general privacy-guideline for the level of downsampling required \cite{ryoo2016lowresegocentric, ryoo2017lowressiamese}.  Starting from $224 \times 224$, we can downsize our images by 16x to $14 \times 14$ images.  We also note that our dataset consists of full-body viewpoints where the face takes up at most $56 \times 56$ of an image; we also perform experiments downsampling by 4x to $56 \times 56$ that provides weaker privacy assurances.    


\textbf{Private Super-resolution: } To perform superresolution, we use a state-of-the-art DCSCN super-resolution network developed by Yamanaka et al \cite{yamanaka2017superresskipconnection}.  The DCSCN is a CNN consisting of a feature extraction network and reconstruction network, and is trained by feeding in pairs of low resolution and high resolution images to find the optimal super-resolution weights.  
We trained two DCSCN models on images downsampled to $56 \times 56$ and $14 \times 14$ on the open-source action recognition dataset NTU-RGDB. 
The NTU RGBD action recognition dataset \cite{shahroudy2016nturgbd}
 consists of 56,880 samples containing RGB and depth map videos of 60 distinct actions recorded using Microsoft Kinect v.2.  We only use the depth data from the dataset to train our DCSCN model.  By training DCSCN models on public datasets disjoint from our dataset, we remove the privacy risk of DCSCN learning from information exclusive to our dataset while still learning a good representation for indoor action depth images.

\section{Datasets}


\textbf{Hand Hygiene Detection: } Significant efforts have been made in developing hand-hygiene compliance technologies to reduce HAIs (hospital acquired infections) \cite{cook2009hai}, which affects a large population of hospital visitors and costs the industry billions of dollars a year \cite{zimlichman2013hai}. Several approaches use RFID and wearable-devices \cite{granadovillar2013handhygiene} with sensor-enabled soap-dispensers to log hand-sanitization. We focus on an visual-based approach which uses depth data and CNNs to identify dispenser usage events \cite{haque2017handhygiene}. Images were collected from an acute care pediatric care unit and an adult ICU from two hospitals. Depth sensors were installed near alcohol-based gel dispensers from top-down and side-view rooms, with a functional range between 0.8 and 4.0 meters. As outlined by \cite{armellino2013handhygiene}, the images are positively labelled when a person correctly follows standard hand-hygiene protocol. A group of ten annotators trained on proper hand hygiene protocol annotated the images, with each depth image analyzed and cross-validated by one to three annotators.The data used for the classifier consists of 113,379 images, of which 11,994 images contained people using the dispenser. We used a train/test split of 90/10.

\begin{figure}[t]
\centering
\includegraphics[width=1.5in]{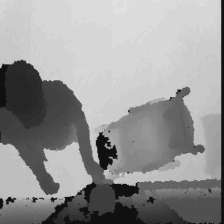}
\includegraphics[width=1.5in]{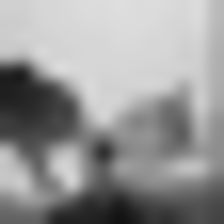}
\includegraphics[width=1.5in]{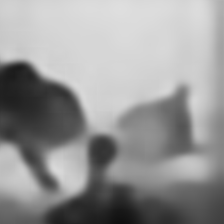}
\caption{From left to right: A high res. depth image, a $16 \times 16$ bicubic downsampled image, and a DCSCN enhanced image.  DCSCN is trained to sharpen the silhouette of the low res. image.}
\label{fig:dcscn_outputs}
\end{figure}

\textbf{ICU Activity Logging:} In ICUs, fine-grained and reliable recording occurences of patient care activities helps improve patient outcomes through enforcing protocol adherence and studying correlation of care activities with care quality \cite{schweickert2009early,team2015early}.  Computer vision-based approaches can help alleviate the workload on nurses and staff through automated activity detection and logging. We collect a dataset of patient care activities in a simulated ICU room.  A depth sensor was used to collect four patient care mobility-related activties: "getting in bed", "sitting on a chair", "getting out of bed", and "standing up from a chair", using a side-view of the room. The simulation was guided by a clinician to ensure the scenarios covered the diverse patient conditions in ICUs; "getting out of bed" for instance may involve different numbers of caretakers and vary significantly in duration.  The data was examined by the clinician to ensure the activities were conducted following correct protocols.
In total, 16853 seconds of videos are collected from 10 actors, comprising of 316 activity instances.
We use 90/10 for our train/test split.  A sample frame from our ICU recordings is included Figure \ref{fig:method}.

\textbf{RGB Images:} We did not obtain RGB videos for our healthcare tasks due to the privacy concerns raised by the participating hospitals and clinicians. We find that depth images are adequate for our experimental tasks, as silhouette information is enough to perform recognition of basic actions like dispenser usage or lying in bed. However, RGB videos are critical for finer-grained activities, especially for tasks involving objects such as surgery or X-ray scans.  Our proposed low-resolution and DCSCN is compatible and perhaps better suited with RGB videos; many low-resolution works use RGB datasets \cite{ryoo2016lowresegocentric, ryoo2017lowressiamese}, and DCSCN techniques were developed primarily for RGB images \cite{zhang2018superresresidualattention, yamanaka2017superresskipconnection}. We believe our proposed technique will provide a method to deploy RGB cameras to hospitals with privacy-preserving assurances, adding more capability to smart-hospital systems.


\section{Experiments}

We perform experiments on both the Hand-Hygiene and ICU tasks with different dimensions and enhancement settings, measuring the accuracy and AUC for each task.

 \begin{table*}[t]
    \centering
    \begin{tabular}{|c|c||c|c|} \hline
    Original Dim & DCSCN & Test Acc. & AUC \\ \hhline{|=|=||=|=|}
    224 $\times$ 224 & No & 94.5\% & 0.987 \\
    \hline
    56 $\times$ 56 & No & 96.27\% & 0.992 \\
    56 $\times$ 56 & Yes & \textbf{98.24\%} & \textbf{0.995} \\
    \hline
    14 $\times$ 14 & No & 92.59\% & 0.9735 \\
    14 $\times$ 14 & Yes & 95.87\% & 0.994 \\
    \hline
    \end{tabular}
    \caption{Hand Hygiene Results: Results from our Resnet-52 model on the hand hygiene task, where we compare results with and without DCSCN enhancement.  Surprisingly, we see that all experiments except unenhanced $14 \times 14$ beat the original performance in terms of both Test Accuracy and AUC.  In addition, we find that DCSCN improves the performance for either both downsampled dimensions.}
        \label{tab:hand_hygiene}
\end{table*}

 \begin{table*}[t]
    \centering
    \begin{tabular}{|c|c||c|c|c|c|c|c|c|} \hline
    Original Dim & DCSCN & Test Acc. & \multicolumn{5}{c|}{AUC} \\ \hline
    Action Class & -- & Average & 0 & 1 & 2 & 3 & 4 \\ \hhline{|=|=||=|=|=|=|=|=|}
    224 $\times$ 224 & No & 68.8\%  & 0.905 & 0.825 & \textbf{0.94} & 0.85 & 0.845 \\
    \hline
    56 $\times$ 56 & No & 70.8\%  & 0.935 & \textbf{0.885} & 0.805 & 0.845 & 0.835\\
    56 $\times$ 56 & Yes & \textbf{72.4\%} & 0.93  & 0.865 & 0.905 & 0.805 & \textbf{0.9} \\
    \hline
    14 $\times$ 14 & No & 66.0\%  &  \textbf{0.97}  & 0.775 & 0.89 & \textbf{0.855} & 0.724 \\
    14 $\times$ 14 & Yes & 62.6\%  & 0.945  & 0.8 & 0.81 & 0.79 & 0.725\\
   \hline
    \end{tabular}
    \caption{ICU Results: Results from our Resnet-18 model on the ICU task, and with columns AUC (n) where n represents the action class of ICU.  We find that the best test accuracy performance is found with DCSCN enhanced $56 \times 56$ images, and that for all of the metrics we find at most 0.12 AUC degradation for all dimensions and at most 4.8\% test accuracy degradation for all dimensions.}
        \label{tab:icu}
\end{table*}



\textbf{Hand Hygiene:} We train an Imagenet-pretrained \cite{Deng09imagenet} Resnet-50 \cite{he2015resnet} model.  We include examples of downsampled images and respective DCSCN outputs in Figure \ref{fig:dcscn_outputs}.  Due to the large class imbalance present in our dataset, we perform data augmentation on our dataset by applying random image transformations and feeding in equal numbers of positive/negative dispenser usage frames during the training phase.  As we can see in Table \ref{tab:hand_hygiene}, scaling down the hand-hygiene dataset does not cause significant drop in utility.  In Table \ref{tab:hand_hygiene}, we find that we actually get the highest performance with the DCSCN-enhanced $56 \times 56$ images at 98.24\%, and higher than baseline results of 94.5\% with DCSCN-enhanced $14 \times 14$ images at 95.87\% which is also slightly better than the performance presented in \cite{haque2017handhygiene}.  Our experiments show that basic downsampled images already preserve a practical amount of utility for each task. For the hand hygiene task, our low resolution outputs even outperform the original images, possibly due to regularization effects that occur with downsampling.  


\textbf{ICU Logging:} We use a Resnet-18 \cite{sherstinsky2018rnnlstm} to process the ICU actions, using data augmentation to balance our classes.  We present our results in Table \ref{tab:icu}, where classes 0-4 represent 'background', 'get in bed', 'get out of bed', 'get in chair', and 'get out of chair'. For the ICU task, we produce comparable or better performance relative to other works \cite{liu2018icu} at 68.8\% test acc.  In Table \ref{tab:icu} we find that scaling down the ICU data also does not cause significant accuracy loss.  
Athough DCSCN does not improve accuracy for $14 \times 14$ images and only slightly improves the accuracy for $56 \times 56$ images, it improves the AUC for several classes such as 2 and 4 for $14 \times 14$ images and class 1 and 4 for $16 \times 16$ images. In addition, Figure \ref{fig:dcscn_outputs} shows that the super-resolution enhanced images are more visually-interpretable and easier to annotate.  Most importantly, we can see how it is visually impossible to discern any personally identifying information from a $14 \times 14$ frame.



\section{Conclusion}
Computer vision in healthcare facilities can greatly aid patient care as seen in tasks like hand hygiene monitoring and ICU logging, but can attract negative sentiment due to the intrusiveness of surveillance systems.  By using downsampled depth images and super-resolution techniques, we can assure a high amount of privacy while preserving enough utility to perform healthcare-relevant action recognition.  Our techniques our compatible with RGB images, and we plan to collect and experiment on a RGB healthcare dataset with state-of-the-art low-resolution action recognition techniques.  We hope the framework we present can promote the development and acceptance of smart-hospitals, and encourage more works to preserve visual privacy in the healthcare domain.

\clearpage
\newpage
\small
\bibliographystyle{abbrv}

\begin{thebibliography}{10}

\bibitem{armellino2013handhygiene}
D.~Armellino, M.~Trivedi, I.~Law, N.~Singh, M.~E. Schilling, E.~Hussain, and
  B.~Farber.
\newblock Replicating changes in hand hygiene in a surgical intensive care unit
  with remote video auditing and feedback.
\newblock American Journal of Infection Control, 2013.

\bibitem{asghar2017hipaaprivacyhealth}
M.~R. Asghar, T.~Lee, M.~M. Baig, E.~Ullah, G.~Russello, and G.~Dobbie.
\newblock A review of privacy and consent management in healthcare: {A} focus
  on emerging data sources.
\newblock {\em CoRR}, abs/1711.00546, 2017.

\bibitem{biswas2006smarthospitalsystem}
J.~Biswas, D.~Zhang, G.~Qiao, V.~Foo, Siang~Fook, Q.~Qiu, and P.~Yap.
\newblock A system for activity monitoring and patient tracking in a smart
  hospital.
\newblock {\em Proceedings of the International Conference on Smart Homes and
  Health Telematics}, 2006.

\bibitem{butler2015privacyutilityrobots}
D.~J. Butler, J.~Huang, F.~Roesner, and M.~Cakmak.
\newblock The privacy-utility tradeoff for remotely teleoperated robots.
\newblock In {\em Proceedings of the Tenth Annual ACM/IEEE International
  Conference on Human-Robot Interaction}, HRI '15, pages 27--34, New York, NY,
  USA, 2015. ACM.

\bibitem{cheng2017facekinectdepth}
Z.~Cheng, T.~Shi, W.~Cui, Y.~Dong, and X.~Fang.
\newblock 3d face recognition based on kinect depth data.
\newblock In {\em 2017 4th International Conference on Systems and Informatics
  (ICSAI)}, pages 555--559, Nov 2017.

\bibitem{cook2009hai}
D.~J. Cook and M.~Schmitter-Edgecombe.
\newblock Assessing the quality of activities in a smart environment.
\newblock {\em Methods of information in medicine}, 48(5):480--485, 2009.

\bibitem{Deng09imagenet}
J.~Deng, W.~Dong, R.~Socher, L.~jia Li, K.~Li, and L.~Fei-fei.
\newblock Imagenet: A large-scale hierarchical image database.
\newblock In {\em In CVPR}, 2009.

\bibitem{zimlichman2013hai}
Z.~E, H.~D, T.~O, and et~al.
\newblock Health care–associated infections: A meta-analysis of costs and
  financial impact on the us health care system.
\newblock {\em JAMA Internal Medicine}, 173(22):2039--2046, 2013.

\bibitem{granadovillar2013handhygiene}
D.~Granado-Villar and B.~Simmonds.
\newblock Utility of an electronic monitoring and reminder system for enhancing
  hand hygiene practices in a pediatric oncology unit.
\newblock 2011.

\bibitem{haque2017handhygiene}
A.~Haque, M.~Guo, A.~Alahi, S.~Yeung, Z.~Luo, A.~Rege, J.~Jopling, N.~L.
  Downing, W.~Beninati, A.~Singh, T.~Platchek, A.~Milstein, and L.~Fei{-}Fei.
\newblock Towards vision-based smart hospitals: {A} system for tracking and
  monitoring hand hygiene compliance.
\newblock {\em CoRR}, abs/1708.00163, 2017.

\bibitem{he2015resnet}
K.~He, X.~Zhang, S.~Ren, and J.~Sun.
\newblock Deep residual learning for image recognition.
\newblock {\em CoRR}, abs/1512.03385, 2015.

\bibitem{hui2018superresinfodistillation}
Z.~Hui, X.~Wang, and X.~Gao.
\newblock Fast and accurate single image super-resolution via information
  distillation network.
\newblock {\em CoRR}, abs/1803.09454, 2018.

\bibitem{keys1981bicubic}
R.~Keys.
\newblock Cubic convolution interpolation for digital image processing.
\newblock {\em IEEE Transactions on Acoustics, Speech, and Signal Processing},
  29(6):1153--1160, December 1981.

\bibitem{li2018lowresfacewild}
P.~Li, L.~Prieto, D.~Mery, and P.~J. Flynn.
\newblock Low resolution face recognition in the wild.
\newblock {\em CoRR}, abs/1805.11529, 2018.

\bibitem{lin2016iotsmarthomeprivacy}
H.~Lin and N.~Bergmann.
\newblock {\em IoT Privacy and Security Challenges for Smart Home
  Environments}, volume~7.
\newblock 07 2016.

\bibitem{liu2018icu}
B.~Liu, M.~G. Guo, E.~Chou, R.~Mehra, S.~Yeung, N.~L. Downing, F.~S. Jopling,
  J.~Jopling, B.~Campbell, K.~Deru, W.~Beninati, A.~Milstein, and L.~Fei-Fei.
\newblock 3d point cloud-based visual prediction of icu mobility care
  activities.
\newblock {\em Machine Learning for Healthcare}, 2018.

\bibitem{luo2018seniorhome}
Z.~Luo, J.-T. Hsieh, N.~Balachandar, S.~Yeung, G.~Pusiol, J.~Luxenberg, G.~Li,
  L.-J. Li, N.~L. Downing, A.~Milstein, and L.~Fei-Fei.
\newblock Computer vision-based descriptive analytics of seniors’ daily
  activities for long-term health monitoring.
\newblock {\em Machine Learning for Healthcare}, 2018.

\bibitem{ma2017icumobility}
A.~Ma, N.~Rawat, A.~Reiter, C.~Shrock, A.~Zhan, A.~Stone, A.~Rabiee,
  S.~Griffin, D.~Needham, and S.~Saria.
\newblock Measuring patient mobility in the icu using a novel noninvasive
  sensor.
\newblock {\em Critical Care Medicine}, 45(4):630--636, 4 2017.

\bibitem{miyazaki2015lowresprivacyconscious}
N.~Miyazaki, K.~Tsuji, M.~Zheng, M.~Nakashima, Y.~Matsuda, and E.~Segawa.
\newblock Privacy-conscious human detection using low-resolution video.
\newblock In {\em 2015 3rd IAPR Asian Conference on Pattern Recognition
  (ACPR)}, pages 326--330, Nov 2015.

\bibitem{noury2008smarthospitaldaynight}
N.~Noury, T.~Hadidi, M.~Laila, A.~Fleury, C.~Villemazet, V.~Rialle, and
  A.~Franco.
\newblock {\em Level of Activity, Night and Day Alternation, and well being
  measured in a Smart Hospital Suite}, volume 2008.
\newblock 02 2008.

\bibitem{perezpellitero2018superrresphotorealistic}
E.~{P{\'e}rez-Pellitero}, M.~S.~M. {Sajjadi}, M.~{Hirsch}, and
  B.~{Sch{\"o}lkopf}.
\newblock {Photorealistic Video Super Resolution}.
\newblock {\em ArXiv e-prints}, July 2018.

\bibitem{ryoo2017lowressiamese}
M.~S. Ryoo, K.~Kim, and H.~J. Yang.
\newblock Extreme low resolution activity recognition with multi-siamese
  embedding learning.
\newblock {\em CoRR}, abs/1708.00999, 2017.

\bibitem{ryoo2016lowresegocentric}
M.~S. Ryoo, B.~Rothrock, and C.~Fleming.
\newblock Privacy-preserving egocentric activity recognition from extreme low
  resolution.
\newblock {\em CoRR}, abs/1604.03196, 2016.

\bibitem{sanchez2008smarthospitalactivityrecognition}
D.~S\'{a}nchez, M.~Tentori, and J.~Favela.
\newblock Activity recognition for the smart hospital.
\newblock {\em IEEE Intelligent Systems}, 23(2):50--57, Mar. 2008.

\bibitem{schweickert2009early}
W.~D. Schweickert, M.~C. Pohlman, A.~S. Pohlman, C.~Nigos, A.~J. Pawlik, C.~L.
  Esbrook, L.~Spears, M.~Miller, M.~Franczyk, D.~Deprizio, G.~A. Schmidt,
  A.~Bowman, R.~Barr, K.~E. McCallister, J.~B. Hall, and J.~P. Kress.
\newblock Early physical and occupational therapy in mechanically ventilated,
  critically ill patients: a randomised controlled trial.
\newblock {\em The Lancet}, 373(9678):1874--1882, 2018/10/28 2009.

\bibitem{shahroudy2016nturgbd}
A.~Shahroudy, J.~Liu, T.~Ng, and G.~Wang.
\newblock {NTU} {RGB+D:} {A} large scale dataset for 3d human activity
  analysis.
\newblock {\em CoRR}, abs/1604.02808, 2016.

\bibitem{sherstinsky2018rnnlstm}
A.~{Sherstinsky}.
\newblock {Fundamentals of Recurrent Neural Network (RNN) and Long Short-Term
  Memory (LSTM) Network}.
\newblock {\em ArXiv e-prints}, Aug. 2018.

\bibitem{team2015early}
S.~Taito, N.~Shime, K.~Ota, and H.~Yasuda.
\newblock Early mobilization of mechanically ventilated patients in the
  intensive care unit.
\newblock {\em Journal of intensive care}, 4:50; 50--50, 07 2016.

\bibitem{twinanda2015smarthospitaldatadriven}
A.~P. Twinanda, E.~O. Alkan, A.~Gangi, M.~de~Mathelin, and N.~Padoy.
\newblock Data-driven spatio-temporal rgbd feature encoding for action
  recognition in operating rooms.
\newblock {\em International Journal of Computer Assisted Radiology and
  Surgery}, 10(6):737--747, Jun 2015.

\bibitem{xu2018lowresactiontwostream}
M.~Xu, A.~Sharghi, X.~Chen, and D.~J. Crandall.
\newblock Fully-coupled two-stream spatiotemporal networks for extremely low
  resolution action recognition.
\newblock {\em CoRR}, abs/1801.03983, 2018.

\bibitem{yamanaka2017superresskipconnection}
J.~Yamanaka, S.~Kuwashima, and T.~Kurita.
\newblock Fast and accurate image super resolution by deep {CNN} with skip
  connection and network in network.
\newblock {\em CoRR}, abs/1707.05425, 2017.

\bibitem{yang2018superresdeeplearning}
W.~{Yang}, X.~{Zhang}, Y.~{Tian}, W.~{Wang}, and J.-H. {Xue}.
\newblock {Deep Learning for Single Image Super-Resolution: A Brief Review}.
\newblock {\em ArXiv e-prints}, Aug. 2018.

\bibitem{yu2018smarthospitaliot}
L.~Yu, Y.~Lu, and X.~Zhu.
\newblock {\em Smart Hospital based on Internet of Things}, volume~7.
\newblock 10 2012.

\bibitem{zangeneh2017lowresfacetwobranch}
E.~Zangeneh, M.~Rahmati, and Y.~Mohsenzadeh.
\newblock Low resolution face recognition using a two-branch deep convolutional
  neural network architecture.
\newblock {\em CoRR}, abs/1706.06247, 2017.

\bibitem{zhang2012privacyfalldetection}
C.~Zhang, Y.~Tian, and E.~Capezuti.
\newblock Privacy preserving automatic fall detection for elderly using rgbd
  cameras.
\newblock In K.~Miesenberger, A.~Karshmer, P.~Penaz, and W.~Zagler, editors,
  {\em Computers Helping People with Special Needs}, pages 625--633. Springer
  Berlin Heidelberg, 2012.

\bibitem{zhang2018superresresidualattention}
Y.~{Zhang}, K.~{Li}, K.~{Li}, L.~{Wang}, B.~{Zhong}, and Y.~{Fu}.
\newblock {Image Super-Resolution Using Very Deep Residual Channel Attention
  Networks}.
\newblock {\em ArXiv e-prints}, July 2018.

\end{thebibliography}

\normalsize
\clearpage
\newpage

\end{document}